\documentclass[11pt,a4paper]{article}
\usepackage[hyperref]{acl2020}
\usepackage{times}
\usepackage{latexsym}

\usepackage{microtype}

\usepackage{booktabs}
\usepackage{multirow}

\aclfinalcopy 

\definecolor{Fern Green}{RGB}{57, 150, 33}
\definecolor{Royal Purple}{rgb}{0.47, 0.32, 0.66}

\title{Practical Perspectives on Quality Estimation for Machine Translation}

\author{Junpei Zhou$^\dag$\Thanks{The work was performed during an internship at Google} \hspace{1.4em} Ciprian Chelba$^\ddag$ \hspace{1.4em} Yuezhang (Music) Li$^\ddag$ \\
  $^\dag$Language Technologies Institute, Carnegie Mellon University \\
  $^\ddag$Google, Inc. \\
  {\tt junpeiz@cs.cmu.edu, \{ciprianchelba,lyzmusic\}@google.com}
}

\date{}

\begin{document}
\maketitle
\begin{abstract}
Sentence level quality estimation (QE) for machine translation (MT) attempts to predict the translation edit rate (TER) cost of post-editing work required to correct MT output. We describe our view on sentence-level QE as dictated by several practical setups encountered in the industry. We find consumers of MT output---whether human or algorithmic ones---to be primarily interested in a binary quality metric: is the translated sentence adequate as-is or does it need post-editing? Motivated by this we propose a quality classification (QC) view on sentence-level QE whereby we focus on maximizing recall at precision above a given threshold. We demonstrate that, while classical QE regression models fare poorly on this task, they can be re-purposed by replacing the output regression layer with a binary classification one, achieving 50-60\% recall at 90\% precision. For a high-quality MT system producing 75-80\% correct translations, this promises a significant reduction in post-editing work indeed.
\end{abstract}

\section{Introduction}

With the development of neural machine translation (NMT) models~\cite{sutskever2014sequence,bahdanau2014neural,vaswani2017attention,edunov2018understanding}, the quality of machine translation systems has been steadily improving over the past few years~\cite{garg2018machine}. 

However, machine translation is still error-prone, producing text that can lack fluency and/or semantic faithfulness to the input. 
Consumers of MT technology often resort to bilingual speakers to post-edit the translated sentences to make them good enough to be used \cite{krings2001repairing}, which is expensive. This option is not available at all to potential algorithmic consumers of MT text output such as a web search engine. Both scenarios create demand for an automatic way to estimate the quality of machine translation output: post-editors could concentrate on low-quality translations and algorithms could filter them out.
This motivates the QE task \cite{specia2010machine,specia2018machine}, which aims to estimate the quality of output from a machine translation system without access to reference translations. 

In this work, we focus on sentence-level QE and describe our exploration and analysis from an industry perspective.
Our contributions are three-fold:
\begin{itemize}
\setlength\itemsep{-0.1em}
    \item We analyze different problem formulations in practical setups encountered in the industry to motivate a binary classification approach to MT QE and introduce the quality classification (QC) task, derived directly from the QE task as defined by WMT.
    \item We adopt a new evaluation metric $R@P_t$ (Recall when Precision is above a threshold $t$) for QC, which is intuitively more meaningful than the MAE/MSE/Pearson metrics used for evaluating QE system performance. The metric is directly correlated with the ratio of translated sentences that are labeled as correct, while controlling for the rate of false-positives.
    \item We conduct experiments for different feature extractors in QC and report recall at different precision thresholds, showing that competitive QE models re-purposed for QC by replacing the output regression layer with a binary classification one could indeed deliver meaningful end-user value, as long as the quality of the underlying MT model is reasonably high.
\end{itemize}

The paper is organized as follows: Section~\ref{sec:relatedwork} describes related work. We motivate the shift to a binary classification setup and associated evaluation metrics in Section~\ref{sec:tasksetting}, followed by a description of our experiments in Section~\ref{sec:exps}, conclusions, and future work directions.

\section{Related Work} \label{sec:relatedwork}

In the traditional MT task setting the evaluation metric is mainly BLEU~\cite{papineni2002bleu}, which compares the translation output with several reference sentences. For QE however, the model tries to predict how far is the translation output by an MT model from its post-edited correct version, without access to reference sentences.

The WMT QE shared task started in 2012~\cite{Callison-Burch:2012:FWS:2393015.2393018} and the most recent one was held in 2019~\cite{fonseca2019findings}.
The goal of the WMT QE sentence-level task is to predict the required post-editing cost, measured in HTER~\cite{snover2006study}, which is a typical regression task.

Among the systems that participated in this shared task, there have been various methods to tackle this problem, and they can be roughly divided into three categories.
The first category uses hand-crafted features, such as those extracted by QuEst++~\cite{specia-paetzold-scarton:2015:ACL-IJCNLP-2015-System-Demonstrations} including sentence length, language model score, and so on.
The second category uses neural models to extract features~\cite{shah-etal-2015-shef,biccici2018rtm}, which encodes a sentence pair into a feature vector.
The third category trains another model as a `teacher', including recent state-of-the-art systems~\cite{kim-lee:2016:N16-1,wang-EtAl:2018:WMT4}. This kind of system is usually composed of two modules: an MT-like source--target encoding model pre-trained with large parallel corpora, stacked with a QE scorer based on the neural features. 
For example, \citet{wang-EtAl:2018:WMT4} adopt the ``Bilingual Expert'' model~\cite{Fan2018BilingualEC} obtained several best results in WMT 2018. \cite{zhou2019source} proposed a model which forces the decoder to attend more to the encoder, instead of being a bi-directional language model. Ensembles of several models as in \cite{kepler2019unbabel} performed best in the 2019 MT QE task.

\section{Task Setting} \label{sec:tasksetting}

\subsection{Classification Instead of Regression} \label{sec:whyclassification}

In the WMT QE sentence-level task, all systems aim at predicting the normalized HTER score for a given \textit{(source, translation)} sentence pair, which is a typical regression task. Submissions are evaluated in terms of the Mean Absolute Error (MAE), Root Mean Squared Error (RMSE), and Pearson's correlation \cite{specia-etal-2018-findings}.

However, after thoroughly analyzing a few user scenarios in the industry, we found that it is more practical to formulate this problem as a classification task instead of a regression task.
There are three primary reasons for this choice:

\paragraph{Simplicity}
The first reason is that most customers don't care too much about the exact (H)TER score, but instead they are interested in knowing which sentences need to be sent for post-editing and which sentences are good enough to be used as such. For example, there is no difference to the user between a translation with 0.7 TER score and another one scoring 0.5 if the TER upper threshold is set to 0.3, because both will be sent for post-editing. 

NMT users are presented with a simple interface: after providing the data we can directly return two sets of sentences to customers; one with text that can be used as-is, and the other with text that needs post-editing.

\paragraph{Understand-ability}
The second reason is that current metrics for WMT QE regression do not give a straightforward sense for how well the model performs.
For example, if we tune the model to make it perform better on low-quality sentences, the MAE and RMSE would decrease.
However, no matter whether the model predicts 0.7 or 0.9 TER for a sentence pair, it would still be sent for post-editing, so the decrease on those regression metrics cannot reflect the real improvement brought to the customers.
What's worse, it is hard to set a threshold for deploying the model in production based on the regression metrics, especially Pearson's correlation. In contrast, by adopting the classification setting, we can use more meaningful metrics such as precision, recall, $F_1$, and so on; see Section \ref{sec:eval} for a detailed discussion on the evaluation metric we choose to work with.

\paragraph{Capability}
The third reason is that training a classifier performs a lot better than setting a threshold on the regression model output when taking the QC view instead of the QE one, see the results in Section~\ref{exps:regression}. 

\subsection{Evaluation Metrics} \label{sec:eval}

There are a lot of metrics available for binary classification, such as accuracy, precision, recall, F-score, confusion matrix, AUC for ROC curve, and so on. After consulting with internal users of NMT technology we decided to use a custom metric $R@P_t$, which maximizes Recall subject to the constraint of Precision being above a threshold $t$. Setting a high value for $P_t$ controls the amount of noise introduced in the downstream pipeline; again, consulting with internal users of NMT the Precision threshold $t = 0.9$ was deemed sufficient; acceptable Recall values depend on task at hand of course.

\section{Experiments} \label{sec:exps}

\subsection{Datasets} \label{sec:dataset}

There are two stages in training the model; the first stage uses a parallel dataset to train the feature extractor (FE) and the second stage uses a QC dataset to train the classifier. The first stage uses the parallel data as described in Section 4.1.1 of \cite{wang2018alibaba} with slightly different pre-processing/filtering. For the QC dataset, we construct a new binary labeled dataset derived from the WMT QE dataset.

QE datasets list source/target sentence pairs with HTER scores as labels; for QC we label samples with 0.0 HTER as `good' (positive) while the rest get `bad` (negative) labels.
In the following, we will denote the converted WMT QE 2017 sentence-level dataset as `WMT17';  Table \ref{tab:datasetinfo} details the number of samples (thousands) in each split and the percentage of positive samples.

\begin{table}[h!]
\centering
    \begin{tabular}{ lccc }
    \toprule
             & \bf Train & \bf Dev & \bf Test\\
    \bf Lang & \multicolumn{3}{c}{\bf Num Samples (Good\%)} \\
    \midrule
    En-De & 23k (14\%)  & 1k (~~9\%)  & 2k (15\%) \\
    De-En & 25k (42\%) & 1k (44\%) & 2k (15\%) \\
    \bottomrule
    \end{tabular}
\caption{WMT17 QE/QC datasets: number of samples (thousands) in each split and the percentage of positive samples in train/dev/test data, respectively.}
\label{tab:datasetinfo}
\end{table}

\begin{table*}[h!]
  \centering
  \begin{tabular}{lccccc}
  \toprule
     &  \multicolumn{2}{c}{\bf En-De} & \multicolumn{2}{c}{\bf De-En} & \\
     \cmidrule{2-5}
    \multicolumn{1}{c}{\bf Hyper-Parameter~~~~~~~~} & \bf Quasi & \bf NMTEx & \bf Quasi & \bf NMTEx & \bf Range \\ 
    \midrule
      qc lstm layers & 1  & 2 & 2 & 2 & [1, 2] \\ 
      qc lstm size & 64 & 256 & 256 & 64 & [64, 128, 256] \\
      qc lstm dropout & 0.0  & 0.2 & 0.0 & 0.1 & [0.0, 0.1, 0.2, 0.3] \\
      learning rate & 1e-5 & 1e-5 & 1e-5 & 1e-5 & [1e-6, 1e-5, 1e-4] \\
    \bottomrule
  \end{tabular}
\caption{Hyper-parameters tuning ranges and the final value adopted by grid-search in the QC model.}
\label{tab:Hyper-parameters}
\end{table*}

\begin{table}[h!]
\centering
    \begin{tabular}{ lcccc }
    \toprule
    \bf Model & \bf Lang & \bf Split & \bf $R@P_{0.8}$ & \bf $R@P_{0.9}$ \\ 
    \midrule
    \multirow{ 4}{*}{Quasi} & En-De & Dev & 0.5111 & 0.4556 \\
     & En-De & Test & 0.4300 & 0.1933 \\
    \cmidrule{2-5}
     & De-En & Dev & 0.7156 & 0.6261 \\
     & De-En & Test & 0.7441 & 0.5255 \\
    \cmidrule{1-5}
    \multirow{ 4}{*}{NMTEx} & En-De & Dev & 0.2111 & 0.0667 \\
     & En-De & Test & 0.2556 & 0.1700 \\
    \cmidrule{2-5}
     & De-En & Dev & 0.7729 & 0.6556 \\
     & De-En & Test & 0.7678 & 0.4904 \\
    \bottomrule
    \end{tabular}
\caption{Experimental results on WMT17 dataset.}
\label{tab:ExperimentalRes}
\end{table}

\subsection{Models}

As described in Section \ref{sec:relatedwork}, the model for MT QE has two components: a feature extractor (FE) and a QE predictor. In regular QE models, the predictor uses an output regression layer and is trained using either MAE or MSE loss. In our QC setup, the output layer is a binary classification one and the training loss is cross-entropy.

We experimented with two options for the FE component. The first is the one described in \cite{zhou2019source}, which got second place in WMT QE 2019 sentence-level task and achieves state-of-the-art among all single models (not ensemble) reported. 
The feature for each token $y_j$ is:
\begin{equation}
    q_j = \mathrm{concat}(\overleftarrow{z_j}, \overrightarrow{z_j}, e^t_{j-1}, e^t_{j+1}, f_j^\mathit{mm})
\end{equation}
where $\overleftarrow{z_j}, \overrightarrow{z_j}$ are state vectors produced by two uni-directional Transformers, and $e^t_{j-1}, e^t_{j+1}$ are embedding vectors of contextual words. 
The $f_j^\mathit{mm}$ is called ``mismatching feature'' as illustrated in \cite{wang2018alibaba}. 
This FE treats the translation system as a black box and trains another model to predict the token based on source tokens and context target tokens; we name it Quasi-MT (denoted `Quasi').

The second FE we experimented with (denoted `NMTEx') is the NMT transformer model as described in Section 3.3 of \cite{caswell-etal-2019-tagged}. The input to the decoder soft-max layer is augmented with the mismatching feature as used in `Quasi' and (optionally) the encoder output before being fed to the QE predictor; the latter was useful for De-En but not for En-De.

FE output is time-reduced using bidirectional LSTM models with dropout and layer normalization as implemented in Lingvo \cite{shen2019lingvo}. The final LSTM state is fed to either a classification layer (QC) or a regression layer (QE). For the QC model training back-propagation gradient is stopped at the underlying FE; dropout in the FE models is set to 0.0 to match it with inference.

\subsection{Experimental Results} \label{sec:experimentalres}

We have implemented the model using Lingvo \cite{shen2019lingvo} for distributed training and conducted experiments in the classification setting described in Section \ref{sec:tasksetting}.
We have tuned hyper-parameters for QC models according to the $R@P_t$ on the development dataset by grid-search, and the final parameters we finally picked are shown in Table \ref{tab:Hyper-parameters}.

The experiments for using NMTEx and Quasi as feature extractors are shown in Table \ref{tab:ExperimentalRes}.
We report results for both $t = 0.8$ and $t = 0.9$ to give a more comprehensive picture of QC model behavior. Results show that competitive QE models re-purposed for classification can attain relatively high (50-60\%) recall at 90\% precision on WMT17 De-En dataset. For an MT system producing 75-80\% correct translations, this would allow labeling 35-50\% of the output sentences as adequate with a small 3-5\% false-positive rate, a significant reduction in post-editing work indeed.

However, we also observe a large discrepancy in QC performance between the En-De and De-En datasets, as well as between En-DE dev and test sets.
The inconsistency between En-De and De-En is possibly due to the fact that they come from different domains relative to the parallel training data: En-De is IT domain and De-En is Pharmaceutical domain.

\subsubsection{Classification versus Regression} \label{exps:regression}

Finally, we conducted experiments on the WMT17 En-De dataset to verify that training a classifier is indeed better than thresholding the output of a regression model. Samples are classified as `good'/`bad' by thresholding the predicted TER output by the QE model described in \cite{zhou2019source}. In a sweep for the threshold value over the [0.0, 0.5] interval on En-De Dev data we could not reach precision higher than 0.64; on En-De Test data there was steep jump from Precision/Recall of 0.32/0.12 to 1.0/0.01. Neither operating points come even close to the ones listed in Table~\ref{tab:ExperimentalRes}.

\section{Conclusions and Future Work}

We have described a practical viewpoint on sentence-level quality estimation for machine translation, as motivated by various scenarios encountered in industry. This leads us to adopt a binary classification framework rather than the regression one used in the WMT QE track. We have described our evaluation metric $R@P_t$, which we find to be more straightforward and meaningful for the end-user of MT output.
We conducted experiments with several feature extractors on data sets derived from the WMT QE ones and showed that competitive QE models re-purposed for classification can attain a 50-60\% recall at 90\% precision.

As for future work, we plan to explore more modeling directions in the classification framework.
We note that in the binary quality setup the classifier output probability is, in fact, indicative of the model's confidence in the translation being either correct or wrong (or uncertain). Besides the shift to confidence scoring, we intend to leverage the fact that in most use cases we have access to both the underlying model that produced the translation and the training data for it.

\bibliography{acl2020}
\bibliographystyle{acl_natbib}

\end{document}